\documentclass{article}

\usepackage{arxiv}

\usepackage[utf8]{inputenc} 
\usepackage[T1]{fontenc}    
\usepackage{hyperref} 
\usepackage{url}
\usepackage{pifont}
\usepackage{graphicx}
\usepackage{mathptmx}
\usepackage{times}
\usepackage{amsmath}
\usepackage{amssymb}
\usepackage{colortbl}
\usepackage{bm}
\usepackage{balance}
\usepackage{algorithm2e}
\usepackage{booktabs}
\usepackage{rotating}
\usepackage{multirow}
\usepackage{amsfonts}       
\usepackage{nicefrac}       
\usepackage{microtype}      
\usepackage{lipsum}

\title{Fully Convolutional Search Heuristic Learning for Rapid Path Planners}

\author{
  Yuka Ariki\\
  Sony Corporation\\ 
  Japan\\
  \texttt{Yuka.Ariki@sony.com} \\
  \And
  Takuya Narihira \\
  Sony Corporation \\
  Japan\\
  \texttt{Takuya.Narihira@sony.com} \\
}
\begin{document}
\maketitle
\begin{abstract}

Path-planning algorithms are an important part of a wide variety of robotic applications, such as mobile robot navigation and robot arm manipulation. However, in large search spaces in which local traps may exist, it remains challenging to reliably find a path while satisfying real-time constraints.
Efforts to speed up the path search have led to the development of many practical path-planning algorithms. These algorithms often define a search heuristic to guide the search towards the goal. The heuristics should be carefully designed for each specific problem to ensure reliability in the various situations encountered in the problem. However, it is often difficult for humans to craft such robust heuristics, and the search performance often degrades under conditions that violate the heuristic assumption.
Rather than manually designing the heuristics, in this work, we propose a learning approach to acquire these search heuristics.
Our method represents the environment containing the obstacles as an image, and this image is fed into fully convolutional neural networks to produce a search heuristic image where every pixel represents a heuristic value (cost-to-go value to a goal) in the form of a vertex of a search graph.
Training the heuristic is performed using previously collected planning results.
Our preliminary experiments (2D grid world navigation experiments) demonstrate significant reduction in the search costs relative to a hand-designed heuristic. 
\end{abstract}

\section{Introduction}
\label{sec:introduction}

Path planning is an important technique for robotic applications such as autonomous mobile robots and arm manipulation. The objective of this technique is to find an optimal or a feasible path from the initial state to the goal, subject to constraints derived from a variety of factors including the non-holonomic property of wheeled mobile robots, collision avoidance, or the limits associated with the joints of a robotic manipulator.

These constraints and the existence of local traps mean that a large amount of calculation is required to find a path.
Traditional search algorithms, A* ~\citep{Hart1968} and Heuristic based Rapidly exploring Random Trees (RRT)~\citep{Vemula2014}, rely on heuristics to decrease the amount of calculations for finding a path.
Typically, these heuristics are manually crafted for a particular robot and environment. For example, to accelerate the path search in three-dimensional (3D) space (i.e., the 2D position and heading angle of the robot) under the non-holonomic constraints of car-like robots, the Hybrid A* algorithm~\citep{Montemerlo2008} combines the following manually designed heuristics: 1) a non-holonomic shortest path cost calculated by Dubins~\citep{Dubins57} or Reeds-and-Shepp~\citep{Reeds1990} algorithms (assuming that no obstacles exist), and 2) a holonomic shortest path cost with obstacles calculated by the backward Dijkstra algorithm. Both require a reasonably smaller amount of computational time compared to the search cost itself. However, it is not trivial for humans to manually craft such heuristics for each specific search problem. Furthermore, although the simple combination of heuristics is effective in simple environments, heuristics do not perform as reliable in more complicated environments. 

\begin{figure*}[ht]
 \begin{center}
   \includegraphics[width=\linewidth]{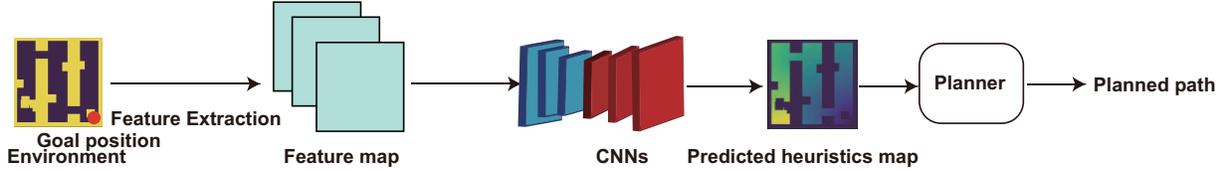}
  \caption{Our fully convolutional system that produces heuristics for planners. The feature maps are extracted from an environment and a goal position, and fed into a CNN. The CNN produces a heuristic map (yellow: high value: distant from the goal, blue: low value: close to the goal), which is used in the path planner module.
  }
\label{fig:procedure}
 \end{center}
\end{figure*}
In recent years, Convolutional Neural Networks (CNNs), a machine-learning technique, have achieved impressive results in a variety of domains such as vision~\citep{Noh2015,Yu2016}, language~\citep{Vinyals2014}, audio~\citep{Hershey2016}, and games~\citep{Silver2017,mnih2015humanlevel} applications. In this paper, we propose CNN-based heuristic learning methods.
As depicted in Figure~\ref{fig:procedure}, our convolutional networks take feature images, which are extracted from an obstacle map, and a goal position as inputs, and predict the heuristic (estimated cost-to-go) values in each position in a 2D map as the outputs. These outputs are used in path planners as heuristics to accelerate the search in planners.
CNNs have the following advantages with respect to learning heuristics in path planning problems, especially for robots; 1) CNNs can capture both the global structures of environments (e.g., the road map) and local details (e.g., the obstacle shape), and 2) CNNs can generate spatially structured outputs (e.g., heuristic values in neighboring configurations tend to be smoothly transitioned).
We utilize path planning algorithms such as Backward Dijkstra and A* to generate ground truth heuristic values. Because our CNNs predict heuristics in a fully convolutional way, both the inference step and training step are efficiently taken on all states in environments at the same time. We also propose a learning method that combines supervision by a planner with the Temporal Difference learning method (TD) to improve sampling efficiency.
Similarly to ours, the approach proposed by Bhardwaj et al. \citep{Bhardwaj2017} learns a heuristic by imitating the Backward Dijkstra algorithm. However, it uses fully connected neural networks, and is applied for each state feature to obtain the state output. This requires the computation of inference and training to be performed independently on each state.
Our model has the ability to learn heuristics only from paths rather than requiring dense cost-to-go values that rely on an algorithm that performs a whole state search, the computational cost of which would be prohibitive for more complicated problems. In addition, our method predicts heuristics for all the states of a 2D map at the same time using fully convolutional NNs, whereas that of~\citep{Bhardwaj2017} predicts a heuristic independently for every state by using fully connected NNs, which is computationally inefficient.
Consequently, they had to employ the DAgger algorithm \citep{Ross2011,Ross2014} to efficiently sample training data. Furthermore, their method relies only on a whole state space search algorithm (i.e., Backward Dijkstra) to generate the ground truth, whereas we propose a more efficient method to generate the ground truth. That is, our method only relies on an optimal path search algorithm (i.e., the A* algorithm). This algorithm enables our method to be applied to problems of a larger scale and a wider range of domains. 

CNNs have previously been employed to solve problems related to path planning.
Wulfmeier et al. ~\citep{Wulfmeier2016,Wulfmeier2015} learned CNNs to produce cost maps from demonstration (i.e., Inverse Reinforcement Learning). The purpose was to learn previously unknown cost functions for planning to imitate the demonstration behavior. Our objective is different in a sense that the heuristic  is learned to reduce the computational time required for planning.
Path planners were previously utilized for reactive CNN policy learning to solve robot navigation problems. For example, \citep{Kanezaki2018} learned a reactive CNN policy with global path planner results in the form of supervised signals. In another study, \citep{Gao2017} utilized global planner paths as inputs to improve a CNN policy based on reinforcement learning.
``Value Iteration Networks (VIN)"~\citep{Tamar2017} embed a differentiable planning module (i.e., value iteration) into CNNs that can learn planners including mapping from observations to cost maps and the state transition probabilities in an end-to-end fashion. \citep{Gupta2017} applied VIN to mobile robot visual navigation problems to perform map localization and planning simultaneously by using an end-to-end framework.
In a VIN framework, the training objective is fundamentally arbitrary, and their experiments show imitation learning and reinforcement learning only, because their objective was not to use learning heuristics to speed up planners.
The computational cost of value iteration becomes prohibitively large when the state space is large. As a result, VIN is limited to search problems with a small state space, e.g., a $28 \times 28$ 2D grid world. Our method does not rely on value iteration at inference time, and can be applied to problems with a much larger search space, e.g., a $224 \times 224$ grid world. Although we limit our experiments to a path-finding problem in simple 2D grid worlds, our method could also be applied to larger problems such as 3D path planning with non-holonomic constraints.
In summary, the main contributions of our paper are as follows:

\begin{itemize}
\item learning heuristics using CNNs in a fully convolutional way over states
\item proposing three learning methods (backward Dijkstra(BD), Sparse, Sparse+TD) which imitate the cost-to-go values generated by either path planning algorithms or the Temporal Difference method
\item demonstrating significant reduction on search costs against a simple heuristic search method in our 2D grid world planning experiments
\end{itemize}

This paper is organized as follows:
in Section~\ref{sec:proposedframework}, we describe the procedure of our proposed framework as illustrated in Figure~\ref{fig:procedure}. First, we describe a search-based path planning algorithm and its heuristic function in Subsection~\ref{subsec:preli}.
Further details of our heuristic learning approach are provided in Subsection ~\ref{subsec:learningh}. We introduce three different algorithms depending on the characteristics of the training data. 
Details of the experiments we performed on these algorithms appear in Section~\ref{sec:ExandR}, where we discuss the results and describe the dataset and present the implementation of CNNs in our proposed framework. These results are used to demonstrate the effectiveness of the proposed framework. Finally, we summarize the results and discuss our future work in Section~\ref{sec:conclusion}.

\section{Proposed Framework}
\label{sec:proposedframework}

\subsection{Preliminaries}
\label{subsec:preli}
We consider a search-based path planner in a graph $G = <V, E>$ as a baseline. The pseudocode of this planner is provided in Algorithm~\ref{alg:graphsearch}. A graph search begins at a start vertex $v_s$. At each vertex evaluation, it expands the next search candidates by $Succ(v)$, which returns successor edges and child vertexes. Each of the search candidate vertexes is validated by the $Valid(e, v, \phi)$ function, which returns $False$ if an edge $e$ is occupied with an obstacle according to an environment $\phi$.
Each valid candidate is evaluated by a search score function $Score(v, \phi)$, and all candidate vertexes are pushed into a queue ${\mathbb O}$ with their scores. At the next iterative cycle, the queue ${\mathbb O}$ pops a vertex with the highest score. Then, its successor vertexes are evaluated and pushed into the queue again. The procedure is repeated until it reaches a goal $v_g$ or the queue ${\mathbb O}$ becomes empty.

\begin{algorithm}
\KwData{\\
  A graph $G=(V,E)$\\
  An environment $\phi$\\
  A start vertex $v_s$ and a goal vertex $v_g$\\
  A search score function $Score: (v, \phi) \rightarrow \mathbb{R}$
}

\BlankLine
\tcc{Starting a search}
${\mathbb O} \leftarrow \emptyset$\;
${\mathbb C} \leftarrow \emptyset$\;

${\mathbb O} \leftarrow {\mathbb O} \cup (Score(v_s, \phi), v_s)$\;

\While {${\mathbb O} \neq \emptyset$} {
  Pop $v$ with the highest score from $\mathbb O$\;
  \If {$v \in {\mathbb C}$} {
    continue\;
  }
  ${\mathbb C} \leftarrow {\mathbb C} \cup v$\;
  \If{$v$ is $v_g$} {
    break\;
  }
  \For {$(v', e) \in {\rm Succ}(v)$} {
    \If {not $Valid(e, v', \phi)$} {
      continue\;
    }
    \If {The cost thus far is larger than any of the previously found paths to $v'$.} {
      continue\;
    }
    $s \leftarrow Score(v', \phi)$\;
    ${\mathbb O} \leftarrow {\mathbb O} \cup (s, v')$\;
    ${\mathbb C} \leftarrow {\mathbb C} \setminus v'$\;
  }
}
Reconstruct a path from the search history\;
\caption{Search-based path planner in a graph}
\label{alg:graphsearch}
\end{algorithm}
Based on Algorithm~\ref{alg:graphsearch}, the Dijkstra search is obtained by defining a score function using a cost-so-far value denoted as $g(v, \phi)$:
\begin{eqnarray}
\label{eq:dijk}
Score(v, \phi) = g(v, \phi)
\end{eqnarray}
A cost-so-far is calculated by accumulating costs (denoted as $Cost(e, v', \phi)$) of edges along a shortest path found so far during a search.
\color{black}
By defining the search heuristics function $h(v, \phi)$, the A* algorithm can be derived from a score function
\begin{eqnarray}
\label{eq:astar}
Score(v, \phi) = g(v, \phi) + h(v, \phi),
\end{eqnarray}
and we define a search depending only on heuristics as a greedy search algorithm as follows:
\begin{eqnarray}
\label{eq:greedy}
Score(v, \phi) = h(v, \phi)
\end{eqnarray}


\subsection{Learning Heuristics using Convolutional Neural Networks for Planner}
\label{subsec:learningh}

Our goal is to find a more efficient heuristic function that minimizes search costs (the number of vertices visited/examined during search). As shown in Figure~\ref{fig:procedure}, our method considers an environment that contains a binary obstacle map as input, extracts the feature maps from it, then uses a CNN to predict a heuristic value at every node $v$ in a graph, which we call heuristic map. The predicted heuristic map is used as a look-up table for querying a heuristic value $h(v, \phi)$ during a graph search based on the planner described in the previous section.
Note that one can extend our method to take a continuous valued cost map as input, where each pixel represents a cost to visit the corresponding state. 

Because the CNN in this method is fully convolutional, it has the ability to simultaneously predict a heuristic value for every node in a graph (single-shot inference). In addition, it can also leverage the matured implementations of GPGPU such as cuDNN.
We learn the heuristic map with the aid of the planner, which is employed during the training of CNNs as a target of prediction. We introduce three variants of learning algorithms.

\begin{figure*}[tb]
 \begin{center}
   \includegraphics[width=15cm]{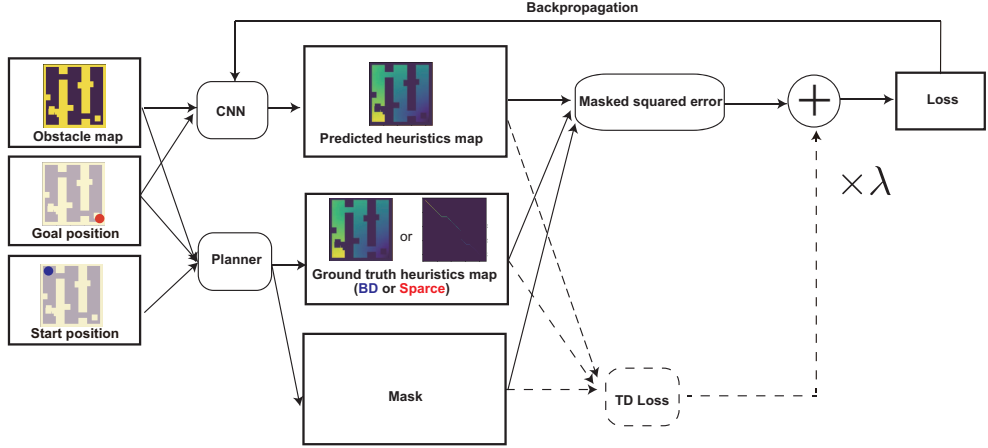}
  \caption{
 Convolutional neural networks (CNNs) are trained to minimize the loss function defined as a masked squared error between the generated cost-to-go from the planner and the current prediction from the CNN. 
 The cost-to-go can be generated either by the Backward Dijkstra (BD) or A* (Sparse). A variant that uses TD error minimization performs iterative value updates to obtain a refined cost-to-go estimation initialized from the current prediction, and composes a loss function combined with the sparse target regression (Sparse+TD). The dashed lines depict Sparse+TD. }
\label{fig:framework}
 \end{center}
\end{figure*}

{\bf Dense target learning with Backward Dijkstra (BD)}:
Our CNN can be directly trained by minimizing the squared error between the prediction and the target cost-to-go value at every node. The cost-to-go of a vertex is defined as the cost accumulated along a shortest path to the goal. The Backward Dijkstra algorithm can calculate the cost-to-go values of all valid vertexes in a graph, where searches are propagated from a goal until no vertex to be opened is available in Algorithm~\ref{alg:graphsearch}. Our training is performed by minimizing the loss function
\begin{eqnarray}
\label{eq:lossdijk}
L(\phi, \hat{H}, M) = \sum_{v \in V} (h(v, \phi) -\hat{H}(v))^2 M(v),
\end{eqnarray}
where $\hat{H}$ denotes the cost-to-go value map generated by the Backward Dijkstra, and $M$ is a mask to make it possible to ignore invalid vertexes the Backward Dijkstra search cannot visit during target value generation, e.g., areas occupied or surrounded by obstacles (Figure.~\ref{fig:framework}).

{\bf Sparse target learning with A* path search (Sparse)}:
The computational time required to generate the cost-to-go target value by the Backward Dijkstra is often prohibitively long for large-scale problems (planning in larger 2D grid maps, or high-dimensional problems, etc.), and can be a bottleneck for learning heuristics. We also propose a learning method that relies only on the target cost-to-go values in those vertexes belonging to the shortest path found by the A* algorithm, given randomly sampled starting and goal positions. A* is much faster than the Backward Dijkstra, which improves the data collection efficiency in terms of the variation of environments. Similar to dense target learning, eq.(\ref{eq:lossdijk}) is used as a loss function, although the training mask $M$ is 1 only at vertexes along a path.

{\bf Sparse target learning with TD error minimization (Sparse+TD)}:
Learning with sparse target signals may result in under-fitting when training due to no supervision signal in pixels that are not visited by the A* path. 
We propose a method to utilize temporal difference (TD) learning method in order to compensate the lack of supervision.

\begin{eqnarray}
\label{eq:td}
\tilde{h}(v, \phi) \leftarrow \min_{(e, v') \in Succ(v)} {Cost(e, v', \phi) + \tilde{h}(v', \phi)},
\end{eqnarray}
where $\tilde{h}(v, \phi)$ is initialized as the current prediction $h(v, \phi)$, $\tilde{h}(v_g, \phi) = 0$, and $\tilde{h}(v \in V_{invalid}, \phi) = \infty$. The value can be updated by using another iterative step and this can be implemented as a convolution with fixed kernels and biases followed by a minimum operation along an axis representing successor vertexes ~\citep{Tamar2017}. Note that we only update the value iteratively during training to obtain more dense target values of the cost-to-go. The use of an updated cost-to-go estimate $\tilde{h}(v)$ enables the loss function to be written as
\begin{eqnarray}
\label{eq:tdloss}
L(\phi, \hat{H}, M) = \sum_{v \in V} \left\{ (h(v, \phi) - \hat{H}(v))^2 M(v) + \lambda (h(v, \phi) - \tilde{h}(v))^2 M_{TD}(v) \right\},
\end{eqnarray}
where $M_{TD}(v)$ is 1 at $v \in \overline{M} \cap V_{valid}$, 0 otherwise, and $\lambda$ balances the weight of the TD minimization loss. We can also update the value iteratively with multiple steps to obtain the target cost-to-go estimate. In our experiment, we used $3$ steps to iteratively update the value and set $\lambda=0.001$. 


\section{Experimental Setup and Results}
\label{sec:ExandR}
\subsection{Dataset}
\label{subsec:dataset}
We trained and evaluated our algorithms on a 2D grid world path planning problem with a dataset provided by~\citep{Bhardwaj2017}.
We used seven different types of environments in the dataset, {\it Shifting gaps}, {\it Bugtrap and Forest}, {\it Forest}, {\it Gap and Forest}, {\it Single Bugtrap}, {\it Mazes}, and {\it Multiple Bugtraps} for our experiment. 
Each environment type contains different kinds of local traps. For example, the environment type {\it Shifting gaps} has an obstacle traversing the central section of the 2D map, obstructing the left and right sides of the map. The traversing obstacle is opened at a vertical position (the position is to be sampled randomly during dataset generation).
Simple heuristics such as Euclidean heuristics may undesirably guide local traps by greedily moving towards the goal without considering the opened position.


Each environment type consists of 800 training 2D grid maps as binary images (either occupied by an obstacle or not), and 100 testing maps. Each map has the dimensionality of $201 \times 201$, where each grid indicates the existence of an obstacle.
We consider each pixel in the map as a vertex and find a path from the start vertex to the goal vertex in an 8-connected grid as a planning problem. The cost is defined as the distance of a path. Edges connected to a vertex at which an obstacle exists are considered as invalid. 
Although we randomly sampled the start and goal positions for path planning to generate supervision during training, we fixed the start and goal positions for path planning as $(0, 0)$ and $(201, 201)$, respectively, during evaluation in order to be compatible with evaluation in~\cite{Bhardwaj2017}. 
\subsection{Implementation details}
\label{subsec:impdetail}
In the neural network architecture we used in our tasks, we employed suitable techniques such as a dilated convolution and encoder-decoder structure to extract global and local spatial contexts from 2D input maps, and to output spatially consistent output images. The encoder CNN repeatedly applied the convolution module three times to produce feature maps with smaller spatial dimensions, and larger maps to take a wider spatial context into account. The convolution module consists of three $3\times 3$ convolutions, and each of them is followed by a batch normalization and a leaky ReLU. A stride of 2 was used in the first convolution, and the dilation factors of the convolution kernels~\citep{Yu2016} were incremented from 1 to 3. The number of convolution channels of the three modules was 16, 32, and 64, respectively. The decoder CNN repeated the deconvolution module three times. The deconvolution module is similar to the convolution module except that the first convolution is replaced with a deconvolution step with a $4\times 4$ kernel with an upscaling factor of 2. The number of convolution channels of the three modules is 32, 16, and 16, respectively, except the last convolution of the third deconvolution module produces single channel output in the form of a heuristics map. 
The input to the CNN consists of feature maps we extract from the 2D obstacle map. The feature maps consist of 1) obstacle map itself, 2) the distances from obstacles, and 3) the distance from the goal, each as an image, which are composed by stacking them as channels of images. The distances from the obstacle and goal are often used to construct Artificial Potential Fields~\cite{Potential2017}, where the goal distance is used as an attractive potential function, and the obstacle distance is used as a repulsive potential function. We pass these functions to train a simple heuristic more easily.

During training, we randomly sample 32 maps from the dataset to construct a mini-batch of a stochastic gradient descent step. For each map, we use A* to randomly sample start and goal positions until a valid path is found between them, after which we generate the cost-to-go targets as described in Section~\ref{subsec:learningh} using either the Backward Dijkstra or A*. A random image translation is applied to inputs as data augmentation, which produces $224 \times 224$ feature maps as inputs.
We used Adam ~\citep{KingmaB14} as a stochastic gradient descent algorithm with $\alpha=0.01$, $\beta_1=0.9$, and $\beta_2=0.999$.
During testing, we used a greedy search algorithm as a planner as in eq.(\ref{eq:greedy}), which accepts the heuristics map produced by our trained CNNs.

\subsection{Results}
\label{subsec:results}
For each type of environment in the dataset, training is performed for $10,000$ epochs. Each training period is approximately 10 hours on a single GTX 1080 Ti for CNNs and on a Core-i7 K7700 for on-the-fly planning ground-truth generation. 

\begin{figure*}[tb]
 \begin{center}
   \includegraphics[width  =\linewidth]{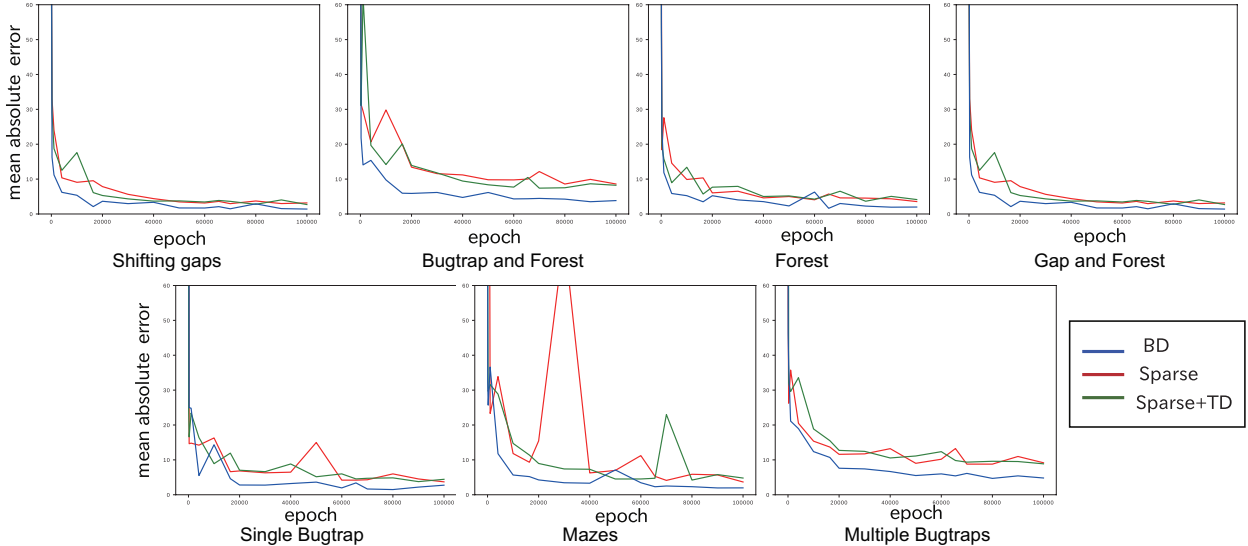}
\caption{Learning curves of prediction values and cost-to-go values obtained by the Backward Dijkstra planner evaluated with mean absolute error. The learning curves of {\it BD}, {\it Sparse}, and {\it Sparse+TD} are plotted with blue, red, and green lines respectively. Each plot represents the curves on different datasets.}
\label{fig:loserror}
 \end{center}
\end{figure*}

Figure~\ref{fig:loserror} shows the training curves of the  mean absolute errors between the predicted heuristic values and ground truth cost-to-go values obtained from the evaluation set. 
{\it BD} consistently produced a smaller error than {\it Sparse} and {\it Sparse+TD} because it can utilize the ground truths on all states in an environment as its training targets. {\it Sparse} and {\it Sparse+TD} produce fairly similar results, although they can only access the ground truths at states along an optimal path.
Table~\ref{tab:comptrainingdata} contains the results of our evaluation of the trained models utilized as heuristic value estimators in a greedy path planner for both metrics: {\it search cost} and {\it path quality}. The {\it search cost} is defined as the number of expanded vertexes during the search, where a smaller number corresponds to a shorter search time.
The {\it path quality} is calculated by accumulating the distance moved along a generated path. 

{\it BD} consistently outperforms the others in both metrics with the learning curve evaluation, whereas no significant difference is observed between {\it Sparse} and {\it Sparse+TD}. We used {\it Sparse} in the subsequent experiments because it maintains a good balance between efficiency in training data generation, algorithm simplicity, and performance. 

\begin{table*}[tb]
    \centering
    \scalebox{0.75}[0.75]{
        \begin{tabular}{c c c c c c c c c c c}
            \toprule
            \midrule
                & &  \multicolumn{3}{c}{Search cost}
                &   \multicolumn{3}{c}{Path quality}\\ 
                \cmidrule{3-5} \cmidrule(l){6-8} 
                &Heuristic& Learned(BD) &Learned(Sparse) & Learned(Sparse+TD)  & Learned(BD)& Learned(Sparse) & Learned(Sparse+TD)  \\ 
                \cmidrule(lr){3-5} \cmidrule(l){6-8}
                \multicolumn{1}{c}{\multirow{7}{*}{\begin{sideways} Environment \end{sideways}}}   &
                \multicolumn{1}{l}{Shifting gap}& 263& 351 & 329 & 322 &350 &353 \\
                \multicolumn{1}{c}{}    &
                \multicolumn{1}{l}{Bugtrap and Forest}&450& 544 & 594   & 359   &  373 & 380\\
                \multicolumn{1}{c}{}    &
                \multicolumn{1}{l}{Forest}& 305& 308 &  465   & 324  &  329 &419  \\
                \multicolumn{1}{c}{}    &   
                \multicolumn{1}{l}{Gaps and Forest}& 300 & 357  & 445  & 337 &348 & 350 \\
                \multicolumn{1}{c}{}    &
                \multicolumn{1}{l}{Single Bugtrap} & 270 & 327 &367   &321 &346 & 342 \\
                \multicolumn{1}{c}{}    &
                \multicolumn{1}{l}{Mazes}  & 401 & 403 & 941 &356 & 370 & 377 \\
                \multicolumn{1}{c}{}    &
                \multicolumn{1}{l}{Multiple Bugtraps} & 855 & 1697 & 1405   &367 & 392 & 376\\
            \midrule
            \bottomrule
        \end{tabular}
        }
        \caption{Comparison of our proposed methods on average search cost and path quality where the following learned heuristics are utilized in the greedy path planner; {\it Learned(BD)} learned convolutional heuristics from the training data produced by the Backward Dijkstra, {\it Learned(Sparse)} by A* (these results are also provided in Table~\ref{tab:allresultsc}), and {\it Learned(Sparse + TD)} from the training data produced by A* and TD error learning }
        \label{tab:comptrainingdata}
\end{table*}
Table~\ref{tab:allresultsc} compares our trained heuristic with a simple Euclid distance heuristic and a heuristic trained with SaIL~\citep{Bhardwaj2017}. 
The results quantitatively show that our method significantly outperforms the other methods in terms of the {\it search cost} in all environments. 
Although the greedy path planner does not aim to find the optimal path, our methods produce paths that closely approximate the optimal path ({\it Optimal}) in terms of the {\it path quality }. {\it Euclid} also produces {\it path quality} close to the ground truth optimal path, because it leads the search aggressively towards the goal, which enables it to find the minimum path length in this simple holonomic 2D path-finding problem. However, its {\it search cost} is far larger than ours. Compared to those of ~\citep{Bhardwaj2017}, our results suggest that our convolutional model predicts more effective heuristics from simply generated feature maps without having the carefully designed features that are fed into simple fully connected networks as in ~\citep{Bhardwaj2017}. Our feature extraction mostly relies on fully convolutional architectures only. 

\begin{figure*}[tb]
 \begin{center}
   \includegraphics[width=15cm]{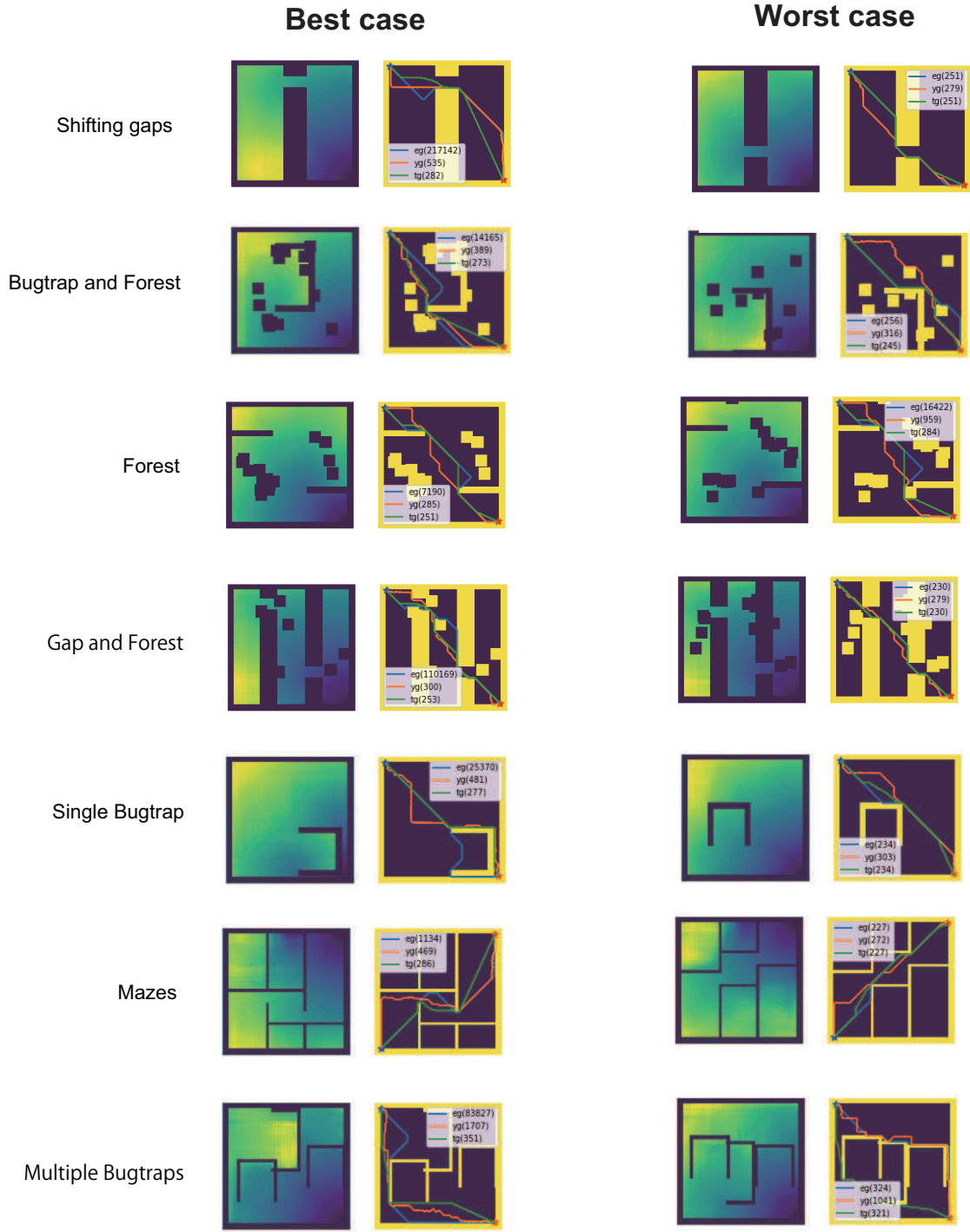}
  \caption{Visualization of the planned paths obtained from the greedy path planner with the following heuristics. {\it Euclid} (blue), {\it Sparse} (orange), and {\it Optimal} (green). On the left, examples for which our methods obtained superior results are shown (best cases), whereas examples for which our methods were inferior are shown on the right (worst cases). For both the best and worst cases, the images on the left-hand side show the predicted heuristics (cost-to-go), where yellow pixels represent higher values (distant from the goal) and blue pixels represent lower values (closer to the goal). Those on the right-hand side show the planned paths. The numbers in the legend on each graph are the search cost. Each row represents a different dataset.}
\label{fig:allresultgoodbad}
 \end{center}
\end{figure*}

We also compared the computational time on our local machine (GTX 1080 Ti and Intel Core-i7 K7700) against the Euclid heuristic baseline. This was not compared with Bhardwaj’s results~\citep{Bhardwaj2017} because their implementation is too slow because of its pure Python implementation. Our planners are written in highly optimized C++, and the CNNs for heuristic prediction utilize GPUs for computation. The average planning time is as follows: 1) A* with Euclidean heuristics: $6.10$ ms, 2) Greedy search with Euclidean heuristics: $4.56$ ms, 3) Greedy search with learned heuristics (ours): $2.48$ ms (CNNs: $2.23$ ms $+$ Planning: $0.25$ ms). The learned heuristics reduces the planning time considerably relative to the Euclidean baseline. Even after adding the computational cost of the CNNs, our method is significantly faster than the baselines.
\begin{table*}[tb]
    \centering
    \scalebox{0.75}[0.75]{
        \begin{tabular}{c c c c c c c c c c c}
            \toprule
            \midrule
                & &  \multicolumn{5}{c}{Search cost}
                &   \multicolumn{4}{c}{Path quality}\\ 
                \cmidrule{3-7} \cmidrule(l){8-11} 
                 &Planner & \multicolumn{3}{c}{Greedy}& A* & SaIL~\cite{Bhardwaj2017}& \multicolumn{3}{c}{Greedy}&SaIL~\cite{Bhardwaj2017} \\ 
                &Heuristic& Optimal &Euclid & Learned(ours)  & Euclid & & Optimal & Euclid & Learned(ours)&\\ 
                \cmidrule(lr){3-7} \cmidrule(l){8-11}
                \multicolumn{1}{c}{\multirow{7}{*}{\begin{sideways} Environment \end{sideways}}}   &
                \multicolumn{1}{l}{Shifting gap}& 250& 37814 & \bf{351} & 23699& \bf{505}& 311&  314 & \bf{350} & \bf{331}  \\
                \multicolumn{1}{c}{}    &
                \multicolumn{1}{l}{Bugtrap and Forest}&273& 20367 & \bf{544}   & 35056& \bf{751} &  325& 352 & \bf{373} & \bf{395}\\
                \multicolumn{1}{c}{}    &
                \multicolumn{1}{l}{Forest}& 252& 9205 &  \bf{308}   & 24418& \bf{357}  &  312 & 334 & \bf{329} &\bf{327} \\
                \multicolumn{1}{c}{}    &   
                \multicolumn{1}{l}{Gaps and Forest}& 259 & 12386  & \bf{357}  & 19981&  \bf{8913}&  316 &  322  &\bf{348} &\bf{945} \\
                \multicolumn{1}{c}{}    &
                \multicolumn{1}{l}{Single Bugtrap} & 236 & 3303 &\bf{327}&27797&\bf{1215}&303&  306  &\bf{346} &\bf{337} \\
                \multicolumn{1}{c}{}    &
                \multicolumn{1}{l}{Mazes}  & 266 & 12687 & \bf{403}  &22013& \bf{1035} & 333 & 337 &  \bf{370} & \bf{428}\\
                \multicolumn{1}{c}{}    &
                \multicolumn{1}{l}{Multiple Bugtraps} & 274 & 19351 & \bf{1697}   &29044& \bf{3182} & 325 & 347 &  \bf{392} & \bf{439}\\
            \midrule
            \bottomrule
        \end{tabular}
        }
        \caption{{\it Search cost} and {\it Path quality} among the following heuristics utilized in the greedy planner: {\it Optimal}: cost-to-go values produced by the Backward Dijkstra, {\it Euclid}: Euclidean distance heuristic, {\it Learned}: convolutional heuristics learned using our proposed framework ({\it Sparse} is used). The results obtained with A* with the Euclidean heuristic and SaIL~\cite{Bhardwaj2017} are also listed for comparison. We used their implementation provided on GitHub to train and test the model using SaIL.}
        \label{tab:allresultsc}
\end{table*}
As shown in Figure~\ref{fig:allresultgoodbad}, the Euclidean heuristic often leads the search to local traps owing to its ignorance of obstacle structures in the environment, which causes undesired search effort. 
One may notice that our model often produces jaggy paths. This effect is attributed to the fact that the heuristic prediction by our CNN model appears globally consistent, but locally noisy. However, our main interest is to reduce the computational cost during path search rather than achieving optimality. Furthermore, we could locally optimize or smooth the obtained feasible paths as a post-processing step.

\section{Conclusion}
\label{sec:conclusion}
In this paper, we proposed a novel CNN-based heuristic learning framework for rapid planners.
Our experiments on path finding problems in 2D grid worlds showed that the proposed learning approaches significantly decrease the search effort compared to a handcrafted heuristic search.
Our convolutional method demonstrated a promising direction to learn the heuristic function to minimize the search cost in complicated environments. One can extend our work to more complicated and high-dimensional search problems, such as non-holonomic path planning problems of mobile robots (e.g., implementing a learning heuristic as CNNs in the Hybrid A* algorithm), and robot arm motion planning (e.g., learning sampling heuristics or the distance metric in RRT).

\clearpage

\end{document}